\documentclass[a4paper, 10pt, conference]{ieeeconf}      % Use this line for a4 paper

\IEEEoverridecommandlockouts                              % This command is only needed if 
\usepackage{graphicx} % for pdf, bitmapped graphics files
\usepackage{amsmath} % assumes amsmath package installed
\usepackage{amssymb}  % assumes amsmath package installed

\usepackage{subcaption}
\usepackage{tikz}
\usepackage{pgfplots}
\pgfplotsset{compat=1.14}

\title{\LARGE \bf
Texture Underfitting for Domain Adaptation
}

\author{Jan-Nico~Zaech$^{1}$\quad Dengxin~Dai$^{1}$\quad Martin~Hahner$^{1}$\quad Luc~Van~Gool$^{1,2}$\\{\tt\small \{zaechj,dai,mhahner,vangool\}@vision.ee.ethz.ch}% <-this % stops a space
\thanks{$^{1}$Computer Vision Laboratory, ETH Zurich, Switzerland}%
\thanks{$^{2}$Dept. of Electrical Engineering ESAT, KU Leuven, Belgium}%
}

\begin{document}

\maketitle
\thispagestyle{empty}
\pagestyle{empty}

%%%%%%%%%%%%%%%%%%%%%%%%%%%%%%%%%%%%%%%%%%%%%%%%%%%%%%%%%%%%%%%%%%%%%%%%%%%%%%%%
\begin{abstract}

Comprehensive semantic segmentation is one of the key components for robust scene understanding and a requirement to enable autonomous driving. Driven by large scale datasets, convolutional neural networks show impressive results on this task. However, a segmentation algorithm generalizing to various scenes and conditions would require an enormously diverse dataset, making the labour intensive data acquisition and labeling process prohibitively expensive. Under the assumption of structural similarities between segmentation maps, domain adaptation promises to resolve this challenge by transferring knowledge from existing, potentially simulated datasets to new environments where no supervision exists. While the performance of this approach is contingent on the concept that neural networks learn a high level understanding of scene structure, recent work suggests that neural networks are biased towards overfitting to texture instead of learning structural and shape information. Considering the ideas underlying semantic segmentation, we employ random image stylization to augment the training dataset and propose a training procedure that facilitates texture underfitting to improve the performance of domain adaptation. In experiments with supervised as well as unsupervised methods for the task of synthetic-to-real domain adaptation, we show that our approach outperforms conventional training methods.

\end{abstract}

%%%%%%%%%%%%%%%%%%%%%%%%%%%%%%%%%%%%%%%%%%%%%%%%%%%%%%%%%%%%%%%%%%%%%%%%%%%%%%%%
\section{INTRODUCTION}
Enabling safe autonomous driving relies on robust scene understanding in a variety of different environments. The encountered variations range from favorable to adverse weather conditions~\cite{SFSU_synthetic}, from daytime to nighttime scenes~\cite{daytime:2:nighttime}, just to name a few. With semantic segmentation being one of the most important methods for scene understanding, substantial work has been invested in improving its robustness over the last decade. On the one hand, modern datasets cover more diverse regions, including cities across Europe in Cityscapes \cite{cordts_cityscapes_2016} and multiple continents as in the Mapillary dataset \cite{neuhold_mapillary_2017}. While extending the variation covered by the datasets is the most intuitive and well understood way to increase the robustness of machine learning algorithms, this approach is limited by the requirement for costly pixel level annotations of images. On the other hand, simulation environments allow for generating high quality renderings of driving scenes together with their pixel level annotations at practically no cost. Even though, recently major progress has been made in making the images appear as realistic as possible, there still remains a domain gap between simulation and reality, resulting in a suboptimal real world performance of algorithms only trained on synthetic data.

Tremendous progress has been made in the last years for domain adaptation. On the one hand, unsupervised domain adaptation aims at overcoming this gap by exploiting similarities such as feature distribution or scene structure between images in the source and target domain. Notable examples include matching data distributions via distance minimization~\cite{adversarial:training:simulated:17,CyCADA} and bridging domain gaps via curriculum domain adaptation~\cite{daytime:2:nighttime,SynRealDataFog19}.
On the other hand, the recipe of supervised fine-tuning has obtained great success for domain adaptation when a limited amount of annotated data is available in the target domain. The goal of this work is to develop a data engineering method such that both unsupervised and supervised domain adaptation methods can benefit from using it.

%With this interpretation and recent publications, indicating 
Inspired by the finding of a recent work~\cite{geirhos_imagenet-trained_2018} that neural networks favor to learn textural rather than structural information, we in this work would like to answer the question whether domain adaptation can benefit from inducing texture under-fitting. We observe that in the context of synthetic-to-real domain adaptation, structures (shapes) are more consistent across the two domains than textures. This implies that a large portion of the domain gap is due to texture differences, thus inducing texture under-fitting for the training of domain adaptation can be beneficial. In order to induce texture under-fitting into the training procedure, this work leverages the success of image stylization to augment the dataset of source domain and target domain with their own stylized versions. The image stylization method is used in a way such that the stylization step randomizes the textures of the images while preserving the structures. 

We have a two-stage training procedure for both unsupervised domain adaptation and supervised domain adaptation: pre-training and fine-tuning. The merit of this simple texture randomization lies in its power of domain generalization. To put in another way, a domain with randomized textures is more general, so a model trained for it is more generalizable to a new domain and two generalized domains have more overlaps than the original domains. This simplifies the training of domain adaptation at the early training stage. Due to this reason, the augmented datasets are better choices than the original source and target datasets for the pre-training. Once the networks are properly initialized via the pre-training, we fine-tune them with conventional data for both domain adaptation scenarios aiming to better adapt to the textures of the target domain. Note that the fine-tuning for unsupervised domain adaptation is performed in an unsupervised manner and the fine-tuning for the supervised domain adaptation is performed in a supervised manner.

Our proposed method is simple, yet effective. We verify its efficacy in two domain adaptation scenarios with extensive experiments. The method is orthogonal to existing domain adaptation methods and can be easily plugged into their training process for further improvement.

\section{RELATED WORK}

\subsection{Semantic Segmentation}
Current methods for semantic segmentation are mostly based on fully convolutional neural networks. After early work introduced the adaptation of classification networks for segmentation tasks \cite{long_fully_2015}, a steady improvement of performance has been archived by capturing contextual information on multiple scales \cite{he_spatial_2014, yu_multi-scale_2015}, extending the receptive field \cite{chen_deeplab_2016} and more powerful backbone networks \cite{he_deep_2016}. Although most state of the art networks are pre-trained on ImageNet \cite{ILSVRC15}, they still require large datasets specific for the segmentation task at hand. 

\subsection{Synthetic Datasets}
The high cost associated with generating annotations for semantic segmentation led to the publication of a large family of synthetic datasets for driving scenes. Inspired by conventional datasets, Richter et al. \cite{leibe_playing_2016, richter_playing_2017} leveraged the consumer market video game Grand Theft Auto V to render images of driving scenes together with their segmentation masks in a semiautomatic manner. Ros et al. \cite{ros_synthia_2016} created an environment specifically for generating the SYNTHIA traffic scene segmentation dataset. By employing rendering methods used in the film industry for the Synscapes dataset \cite{wrenninge_synscapes:_2018}, Wrenninge et al. recently made a leap forward in generating realistically looking images. There are also works creating semi-synthetic data for adverse driving conditions by imposing weathering effects into real images \cite{SFSU_synthetic}. Orthogonal to this directions, the CARLA urban driving simulator \cite{pmlr-v78-dosovitskiy17a} offers the possibility of interaction between the algorithm and the environment together with real time image rendering. There is an emerging stream of methods in synthesizing weather effects into normal images to create weather-degraded images for semantic scene understanding under adverse weather conditions~\cite{SFSU_synthetic,NighttimeSegmentation19,SynRealDataFog19}. Although substantial work has been invested in providing a realistic image appearance, the domain gap between simulated and real data still exists, which evokes the need for domain adaptation methods.

\subsection{Domain Adaptation}
Domain adaptation aims at overcoming performance degradation when the data distribution at test time does not match the distribution present during training. Early publications on domain adaptation in classical machine learning \cite{pan_survey_2010} as well as in deep learning \cite{wang_deep_2018} mainly addressed domain mismatch in classification tasks. In this context, the alignment of source and target distribution in the feature space has been proven to work well \cite{Ganin:2015:UDA:3045118.3045244,tzeng_simultaneous_2015,ganin_domain-adversarial_2015}. In addition to this, domain adaptation for semantic segmentation of road scenes can profit from similar scene structures between domains~\cite{curriculum:domain:adaptation:17,chen2018road} -- while the appearance or texture might change significantly from source to target domain, the overall spatial layouts are similar. Tsai et at. \cite{tsai_learning_2018} employed an adversarial learning scheme leveraging this property and achieved a significant performance improvement over previous methods. Recently, this work was extended to also leverage the more powerful local similarities by patch matching \cite{tsai_domain_2019}. Curriculum Model Adaptation adapts models from an easier task to a harder task in a step-by-step fashion and has achieved great performance in multiple domain adaptation scenarios~\cite{curriculum:domain:adaptation:17,daytime:2:nighttime,SynRealDataFog19,NighttimeSegmentation19}.

\subsection{Texture Bias}
Interpretation of convolutional neural network features is an open and active field of research. While the common belief that networks learn a fine to coarse structural representation of objects present during training is supported by network visualization techniques and widely accepted in the field \cite{kriegeskorte_deep_2015,lecun_deep_2015}, recent work suggests that networks are heavily relying on textural information \cite{geirhos_imagenet-trained_2018,gatys_texture_2017,brendel_approximating_2019}. By restricting the receptive field of neural networks, Brendel et al. \cite{brendel_approximating_2019} were able to show that ImageNet classification performance does not drop significantly, even when only relying on textural information. Furthermore, extensive user studies by Geirhos et al. \cite{geirhos_imagenet-trained_2018} indicate that in contrast to humans, neural networks are biased towards performing inference based on texture rather than on the structure of images. By stylizing the ImageNet training dataset, they were able to reduce the texture bias and made the network behave more like human test subjects on classification tasks. In contrast to this, we investigate the influence of texture bias in domain adaptation for segmentation tasks, where, compared to classification, much more local information, as well as information on different scales is required. Furthermore, we focus on the evaluation of different network training approaches among themselves, instead of showing a general difference between human intuition and network prediction.

\section{APPROACH}

\subsection{Overview}
\begin{figure}[tb]
    \centering
    \begin{subfigure}[t]{.49\linewidth}
        \centering
        \includegraphics[width=\linewidth]{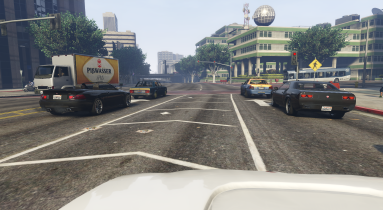}
    \end{subfigure}
    \begin{subfigure}[t]{.49\linewidth}
        \centering
        \includegraphics[width=\linewidth]{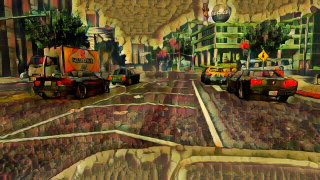}
        \vspace{0cm}
    \end{subfigure}
    \begin{subfigure}[t]{.49\linewidth}
        \centering
        \includegraphics[width=\linewidth]{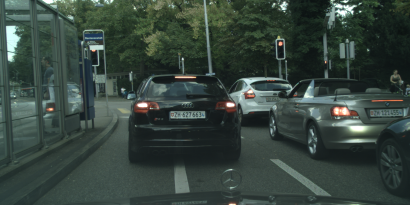}
    \end{subfigure}
    \begin{subfigure}[t]{.49\linewidth}
        \centering
        \includegraphics[width=\linewidth]{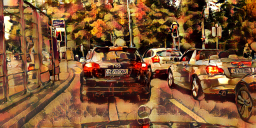}
    \end{subfigure}
    \caption{Examples images from the original and stylized Cityscapes and Playing for Data dataset. On the left, the original image is visualized. The corresponding result after stylization via AdaIN \cite{huang_arbitrary_2017} is shown on the right.}
    \label{fig:stylization_comp}
\end{figure}

Our domain adaptation approach for semantic segmentation addresses the problem of overcoming textural differences between source and target domain by introducing a scene structure dataset into the training procedure. The scene structure dataset is generated by stylizing the original images with the style from random paintings as proposed in \cite{geirhos_imagenet-trained_2018} and called stylized dataset in the following. A comparison of images from the Cityscapes dataset as well as from Playing for Data together with the corresponding stylized image is shown in Figure \ref{fig:stylization_comp}.

Using random paintings for style transfer ensures that the texture of the same object class varies between images and hence does not encode information about it. As a result, the assignment of a class labels rely more on object shapes which is less domain-dependent than object textures. Object recognition with only shape information seems to be an inherently harder task. However, a human observer can still easily recognize and annotate almost all objects in the scene based purely on shape information. This recognition requires information from a more global context, and enforcing algorithms to learn this more global shape information which is more transferable across domains. 

\subsection{Stylization}
For training and evaluation, adaptation from Playing for Data (GTA) \cite{leibe_playing_2016} to Cityscapes (CS) \cite{cordts_cityscapes_2016} is considered. Playing for Data is a dataset consisting of 24966 synthetic trafic scene images rendered with the Grand Theft Auto V engine. Cityscapes is a driving datasets consisting of 3475 publicly available images with pixel level segmentation for training and validation acquired in 50 European cities. To generate a stylized version of each dataset, we perform feed-forward style transfer with adaptive instance normalization \cite{huang_arbitrary_2017}. Images from the driving datasets are used as content images and each is transformed by adopting a style from paintings in the Painter by Numbers dataset hosted on Kaggle\footnote{https://www.kaggle.com/c/painter-by-numbers}. As the Painter by Number dataset with 79434 paintings is larger than Playing for Data as well as Cityscapes with 24966 and 3475 annotated training and validation images respectively, each image can be stylized differently.

With the additional stylized datasets, multiple combinations are possible during training. In the following we denote the dataset combinations as:

\begin{flushleft}
\begin{tabular}{@{}ll@{}}
(GTA + CS)\,: & Conventional dataset \\
stylized(GTA + CS)\,: & Stylized dataset \\
stylized(GTA + CS) + (GTA + CS)\,: & Combined dataset \\
\end{tabular}
\end{flushleft}

\subsection{Unsupervised Domain Adaptation}
\begin{figure}
    \centering
    \includegraphics[width=\linewidth]{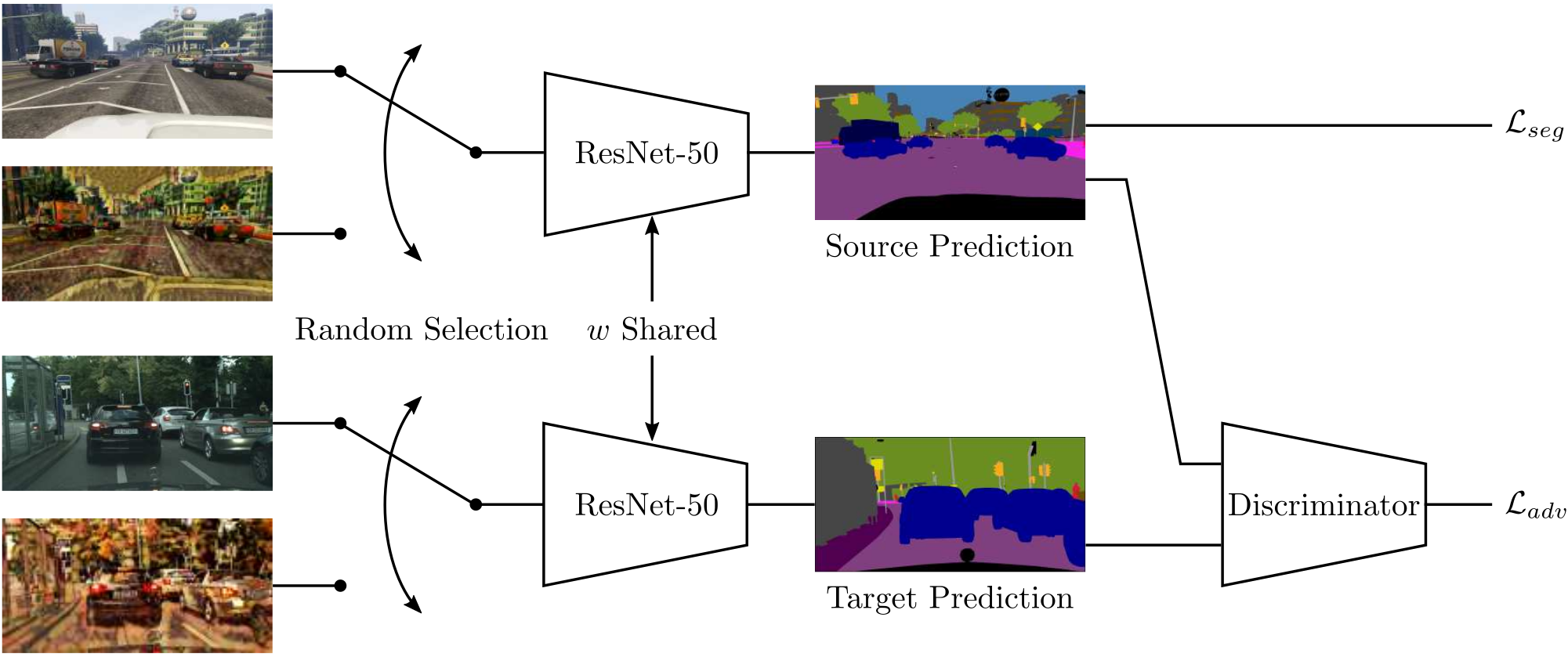}
    \caption{Unsupervised domain adaptation pipeline.}
    \label{fig:network_architecture}
\end{figure}

We base our unsupervised domain adaptation network on the AdaptSegNet structure as proposed by Tsai et al. \cite{tsai_learning_2018}. For further details we refer the reader to \cite{tsai_learning_2018}, where our pipeline is adapted from. For segmentation, the DeepLab-v2 architecture \cite{chen_deeplab_2016} with ResNet-50 backbone is used, which reduces training time and allows for a statistical comparison between different training approaches. We extend the data loader to use the combined dataset via a random dataset selection approach, as shown in Equation \ref{eq:dataset structure}, where the input images are drawn with equal probability either from the conventional or from the stylized dataset. The full architecture is visualized in Figure~\ref{fig:network_architecture}.

\begin{equation}
\begin{aligned}
&\text{Conventional:}\\
&(X_{\text{GTA}}, Y_{\text{GTA}}, X_{\text{CS}}) &&|\,X \in \mathbf X_{\text{conventional}}\\[5pt]
&\text{Stylized:}\\
&(X_{\text{GTA}}, Y_{\text{GTA}}, X_{\text{CS}}) &&|\,X \in \mathbf X_{\text{stylized}}\\[5pt]
&\text{Combined:}\\
&(X_{\text{GTA}}, Y_{\text{GTA}}, X_{\text{CS}}) &&|\,X \in \left\{
\begin{array}{ll}
     \mathbf X_{\text{conventional}} &p=0.5 \\
     \mathbf X_{\text{stylized}} &p=0.5
\end{array} \right.\\
\end{aligned}
\label{eq:dataset structure}
\end{equation}

While inducing a strong structural bias during network training is intuitively sensible for global tasks like classification \cite{geirhos_imagenet-trained_2018}, semantic segmentation requires additional local information to infer a dense mask. Thus, direct domain adaptation from the combined dataset is a much harder task than classical domain adaptation, resulting in performance degradation, even though it better captures the underlying idea. To overcome these challenges and unify local as well as global information in domain transfer, we split training into two stages. In the pre-training stage training is performed on the combined dataset and terminated after a predefined number of iterations. In the subsequent fine-tuning stage, training is continued with the conventional dataset. With this approach, pre-training prevents the network from overfitting to texture, while simultaneously reducing the domain gap with the stylized images from both datasets. However, as the task of segmentation is much more challenging with stylized images, the resulting segmentation masks in the source domain, used for training the disciminator network do not reach a sufficient quality. Therefore, fine-tuning, which only uses the conventional dataset, allows the network to refine the prediction and pick up textural cues that are required for providing he final high-quality segmentation masks.

\subsection{Supervised Domain Adaptation with Limited Data}
Supervised domain adaptation with a limited amount of labeled data is a second important approach towards successfully domain adaptation. While labeling a dataset consisting of thousands of images is too costly in many scenarios, it is generally feasible to annotate a handful of images from the target domain. With those images, a network pre-trained on a related, larger dataset can be fine-tuned, resulting in a better performance in the target domain without overfitting to the small dataset.

In our experiments, we run pre-training with cross entropy loss on either GTA, stylized GTA or combined GTA for 60,000 iterations with all 24966 synthetic images in the source domain. For combined GTA in each iteration an image is randomly chosen either from the conventional or stylized Playing for Data dataset with probability $p=0.5$. Subsequently, training is resumed with the small fine-tuning dataset of 5, 10, and 20 images selected from the original Cityscapes training set (target domain) for 10,000 additional iterations. By performing pre-training on the combined dataset, the network is forced to learn structural information in addition to the textural cues, allowing for a more efficient domain transfer.

\section{EXPERIMENTS AND RESULTS}

\begin{table*}[tb]
\centering
%\resizebox{\textwidth}{!}{\begin{tabular}{l||l|l|l|l|l|l|l|l|l|l|l|l|l|l|l|l|l|l|l||l}
\resizebox{\textwidth}{!}{\begin{tabular}{c||c|c|c|c|c|c|c|c|c|c|c|c|c|c|c|c|c|c|c||c}
\textit{Training}
&  \rotatebox[origin=c]{90}{road}   
&  \rotatebox[origin=c]{90}{sidewalk} 
&  \rotatebox[origin=c]{90}{building} 
&  \rotatebox[origin=c]{90}{wall}   
&  \rotatebox[origin=c]{90}{fence}   
&  \rotatebox[origin=c]{90}{pole}   
&  \rotatebox[origin=c]{90}{tr.light} 
&  \rotatebox[origin=c]{90}{tr.sign} 
&  \rotatebox[origin=c]{90}{vegetation} 
&  \rotatebox[origin=c]{90}{terrain} 
&  \rotatebox[origin=c]{90}{sky}   
&  \rotatebox[origin=c]{90}{person} 
&  \rotatebox[origin=c]{90}{rider}  
&  \rotatebox[origin=c]{90}{car}    
&  \rotatebox[origin=c]{90}{truck}  
&  \rotatebox[origin=c]{90}{bus}    
&  \rotatebox[origin=c]{90}{train}    
&  \rotatebox[origin=c]{90}{motorcycle}
&  \rotatebox[origin=c]{90}{bicycle} 
&  \rotatebox[origin=c]{90}{\textbf{mean IoU}} \\[20pt]
               \hline 

\cline{1-21}

No Adaptation
&42.17
&15.03
&64.84
&16.05
&\textbf{12.72}
&\textbf{21.92}
&16.96
&14.25
&72.60
&17.69
&56.70
&34.43
&6.33
&18.82
&9.72
&4.13
&\textbf{8.26}
&8.83
&0.98
&23.29
\\

\hline
Conventional Dataset \cite{tsai_learning_2018}
&81.31
&21.98
&73.16
&\textbf{17.99}
&10.83
&20.68
&24.25
&15.05
&78.55
&\textbf{24.86}
&68.48
&\textbf{46.20}
&14.54
&60.32
&20.39
&15.30
&4.45
&\textbf{17.00}
&8.54
&32.84\\
\hline

Combined Dataset
&\textbf{84.62}
&\textbf{23.57}
&\textbf{75.27}
&17.58
&12.03
&21.05
&\textbf{24.84}
&\textbf{16.25}
&\textbf{78.96}
&21.00
&\textbf{72.70}
&44.14
&\textbf{15.36}
&\textbf{69.08}
&\textbf{22.34}
&\textbf{16.74}
&1.62
&16.73
&\textbf{10.80}
&\textbf{33.93}\\
\hline

\hline

\end{tabular}}
\caption{Results on the Cityscapes Validation set with unsupervised domain adaptation, evaluated over 10 runs and a window of 6 iterations. Results without adaptation from training terminated after 60000 iterations.}
\label{table:unsupervised results}
\end{table*}

\begin{table*}[tb]
\centering
%\resizebox{\textwidth}{!}{\begin{tabular}{l||l|l|l|l|l|l|l|l|l|l|l|l|l|l|l|l|l|l|l||l}
\resizebox{\textwidth}{!}{\begin{tabular}{c||c|c|c|c|c|c|c|c|c|c|c|c|c|c|c|c|c|c|c||c}
\textit{Training}
&  \rotatebox[origin=c]{90}{road}   
&  \rotatebox[origin=c]{90}{sidewalk} 
&  \rotatebox[origin=c]{90}{building} 
&  \rotatebox[origin=c]{90}{wall}   
&  \rotatebox[origin=c]{90}{fence}   
&  \rotatebox[origin=c]{90}{pole}   
&  \rotatebox[origin=c]{90}{tr.light} 
&  \rotatebox[origin=c]{90}{tr.sign} 
&  \rotatebox[origin=c]{90}{vegetation} 
&  \rotatebox[origin=c]{90}{terrain} 
&  \rotatebox[origin=c]{90}{sky}   
&  \rotatebox[origin=c]{90}{person} 
&  \rotatebox[origin=c]{90}{rider}  
&  \rotatebox[origin=c]{90}{car}    
&  \rotatebox[origin=c]{90}{truck}  
&  \rotatebox[origin=c]{90}{bus}    
&  \rotatebox[origin=c]{90}{train}    
&  \rotatebox[origin=c]{90}{motorcycle}
&  \rotatebox[origin=c]{90}{bicycle} 
&  \rotatebox[origin=c]{90}{\textbf{mean IoU}} \\[20pt]
\hline

\hline

\multicolumn{21}{c}{Cityscapes 5}\\
\hline

\cline{1-21}

GTA
&89.56
&40.14
&77.17
&\textbf{18.25}
&9.06
&25.97
&21.59
&19.37
&80.88
&34.46
&58.8
&49.24
&4.55
&77.89
&\textbf{28.51
}&24.83
&\textbf{16.19}
&\textbf{17.96}
&\textbf{34.70}
&38.38\\
\hline

SGTA
&89.15
&36.62
&77.35
&14.60
&\textbf{11.73}
&17.51
&11.63
&7.17
&76.60
&20.34
&49.73
&44.22
&9.99
&74.79
&28.37
&32.82
&3.81
&8.14
&1.50
&32.43\\
\hline

CGTA
&\textbf{90.44}
&\textbf{42.19}
&\textbf{80.56}
&16.37
&10.58
&\textbf{28.75}
&\textbf{21.85}
&\textbf{26.48}
&\textbf{82.70}
&\textbf{35.43}
&\textbf{85.29}
&\textbf{49.43}
&\textbf{10.35}
&\textbf{78.51}
&24.26
&\textbf{33.72}
&1.12
&11.61
&23.77
&\textbf{39.65}\\
\hline

\hline

\multicolumn{21}{c}{Cityscapes 10}\\
\hline

\cline{1-21}

GTA
&90.87
&39.48
&79.39
&14.37
&6.77
&29.06
&\textbf{24.31}
&26.69
&82.99
&30.27
&72.27
&\textbf{47.58}
&5.30
&\textbf{79.52}
&\textbf{30.27}
&25.81
&\textbf{17.69}
&\textbf{18.23}
&\textbf{38.36}
&39.96\\
\hline

SGTA
&90.61
&43.56
&78.35
&12.00
&7.41
&24.18
&16.00
&9.48
&80.13
&28.36
&54.56
&45.36
&\textbf{8.64}
&76.20
&26.80
&31.44
&2.47
&8.65
&9.88
&34.43\\
\hline

CGTA
&\textbf{91.54}
&\textbf{46.66}
&\textbf{81.29}
&\textbf{16.34}
&\textbf{11.54}
&\textbf{29.88}
&23.12
&\textbf{27.78}
&\textbf{83.59}
&\textbf{32.77}
&\textbf{82.19}
&47.26
&5.82
&79.04
&25.16
&\textbf{32.36}
&0.54
&15.4
&36.68
&\textbf{40.47}\\
\hline

\hline

\multicolumn{21}{c}{Cityscapes 20}\\
\hline

\cline{1-21}

GTA
&91.26
&46.13
&80.80
&\textbf{16.59}
&21.58
&\textbf{30.26}
&22.87
&25.99
&83.93
&33.06
&72.54
&\textbf{46.69}
&\textbf{17.76}
&\textbf{79.95}
&\textbf{29.97}
&30.37
&\textbf{8.01}
&\textbf{18.38}
&\textbf{40.06}
&41.91\\
\hline

SGTA
&91.54
&48.24
&79.44
&13.00
&21.59
&26.08
&17.18
&13.76
&82.34
&32.16
&61.97
&43.18
&15.81
&77.78
&27.10
&32.39
&2.47
&10.82
&36.47
&38.60\\
\hline

CGTA
&\textbf{92.09}
&\textbf{50.72}
&\textbf{81.87}
&16.55
&\textbf{23.24}
&30.20
&\textbf{23.21}
&\textbf{26.51}
&\textbf{84.57}
&\textbf{34.42}
&\textbf{83.09}
&45.21
&17.13
&79.23
&27.28
&\textbf{35.77}
&0.01
&9.15
&36.25
&\textbf{41.92}\\
\hline

\hline

\end{tabular}}
\caption{Results on the Cityscapes validation set with supervised domain adaptation on 5, 10, and 20 training images. Checkpoints selected with the best performance on 100 images held back during training.}
\label{table:supervised results}
\end{table*}

\begin{figure}
\centering
\begin{tikzpicture}
\begin{axis}[
    title={Average mIoU on the Cityscapes validation set},
    xlabel={Iteration},
    ylabel={mIoU [\%]},
    xmin=60000, xmax=160000,
    ymin=30, ymax=35,
    xtick={60000,80000,100000,120000,140000,160000},
    ytick={30,31,32,33,34,35},
    legend pos= south east ,
    ymajorgrids=true,
    grid style=dashed,
    ylabel near ticks,
]

\addplot[
    color=blue,
    mark=+,
    ]
    coordinates {
    (5000,23.791)(10000,24.98)(15000,27.524)(20000,28.478)(25000,27.968)(30000,29.88)(35000,29.798)(40000,29.644)(45000,31.112)(50000,31.183)(55000,30.355)(60000,31.029)(65000,31.48)(70000,31.162)(75000,30.734)(80000,32.535)(85000,31.929)(90000,32.35)(95000,31.643)(100000,32.19)(105000,32.979)(110000,32.706)(115000,33.18)(120000,32.904)(125000,32.57)(130000,32.664)(135000,32.992)(140000,32.023)(145000,31.65)(150000,32.52)(155000,32.422)(160000,32.154)
    };
    
\addplot[
    color=red,
    mark=+,
    ]
    coordinates {
    (85000,26.13)(90000,33.177)(95000,33.144)(100000,34.2)(105000,32.287)(110000,33.514)(115000,33.05)(120000,33.599)(125000,33.988)(130000,33.87)(135000,33.949)(140000,33.899)(145000,33.685)(150000,34.204)(155000,33.406)(160000,34.154)
    };

\legend{Conventional dataset ,Combined datased}
    
\addplot[
    color=blue,
    dashed,
    ]
    coordinates {
    (0,33.18)(160000,33.18)
    };

\end{axis}
\end{tikzpicture}
\caption{Average mIoU on the Cityscapes validation set. Plots are averaged over ten runs for each setup.}
\label{fig:unsupervised performance}
\end{figure}
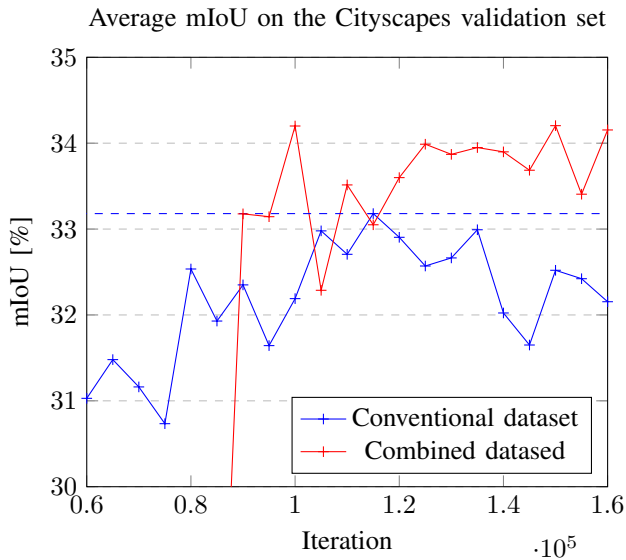

\begin{figure}[!h]
    \centering
    \begin{subfigure}[t]{\linewidth}
        \centering
        \begin{subfigure}[t]{.49\linewidth}
            \centering
            \includegraphics[width=\linewidth]{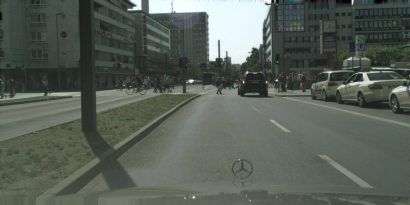}
        \end{subfigure}
        \begin{subfigure}[t]{.49\linewidth}
            \centering
            \includegraphics[width=\linewidth]{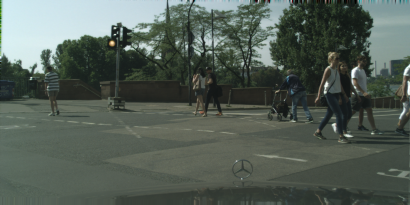}
        \end{subfigure}
        \caption{Input image.}
        \vspace{5pt}
    \end{subfigure}
        \begin{subfigure}[t]{\linewidth}
        \centering
        \begin{subfigure}[t]{.49\linewidth}
            \centering
            \includegraphics[width=\linewidth]{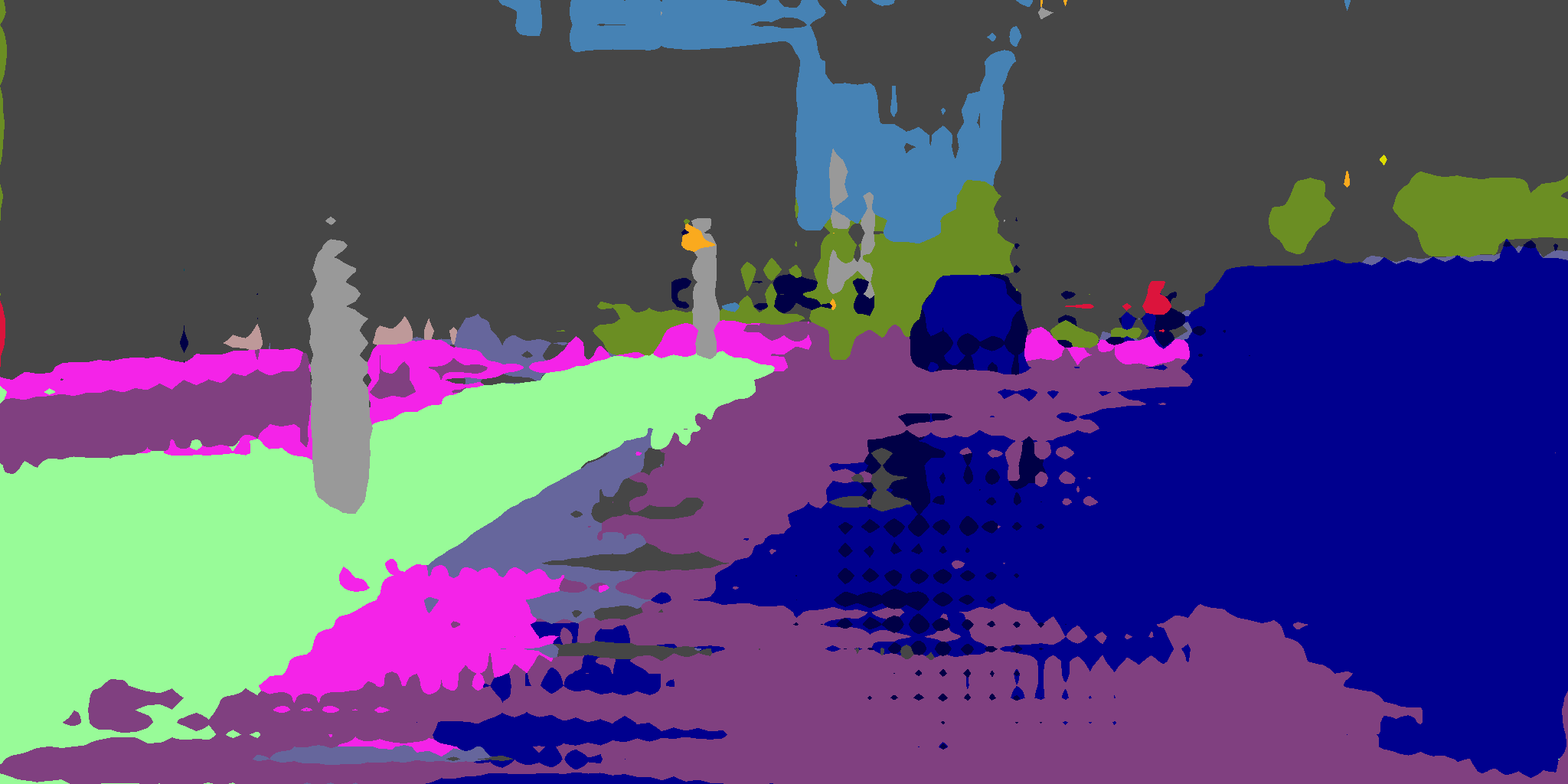}
        \end{subfigure}
        \begin{subfigure}[t]{.49\linewidth}
            \centering
            \includegraphics[width=\linewidth]{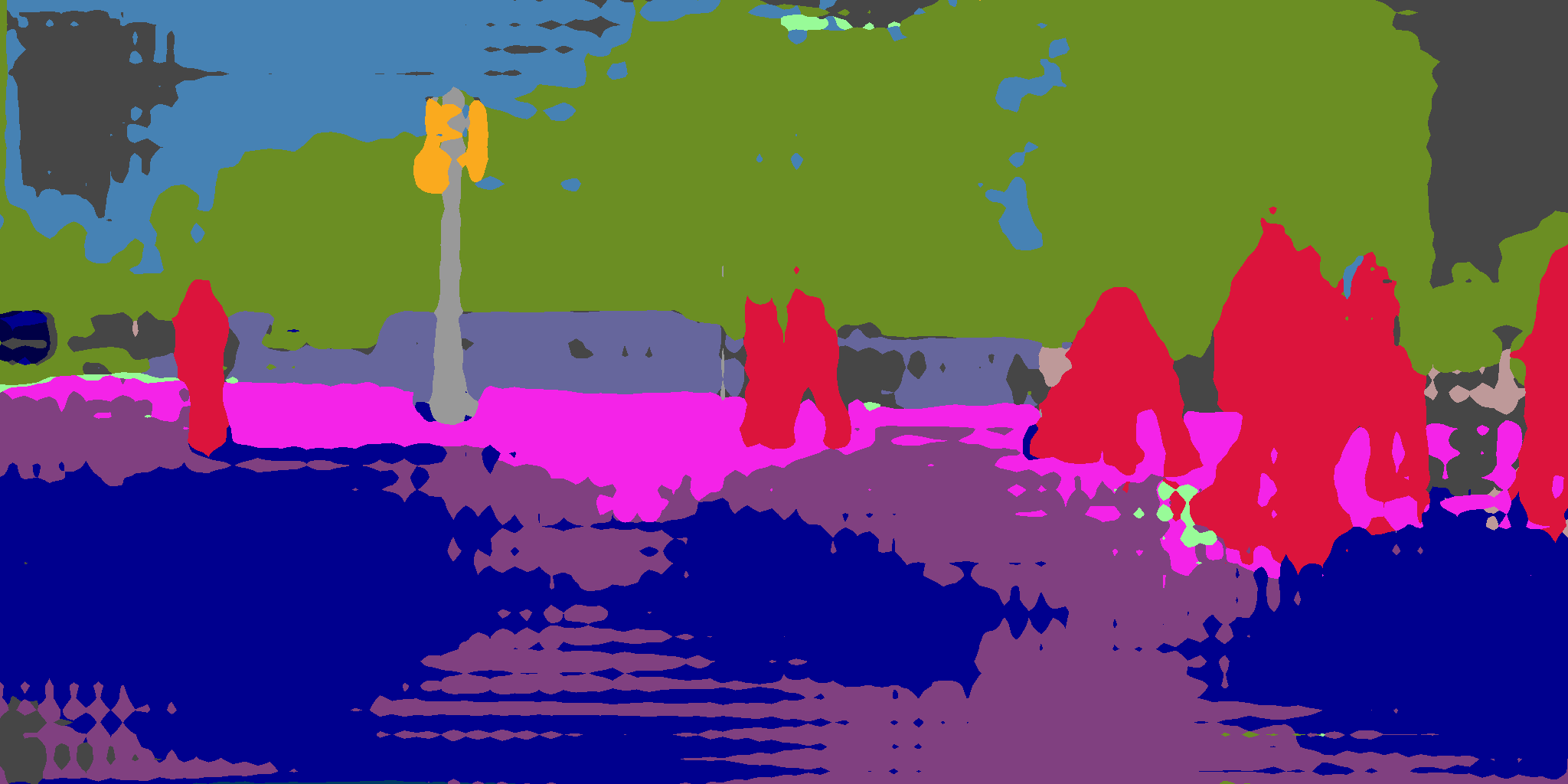}
        \end{subfigure}
        \caption{No adaptation.}
        \vspace{5pt}
    \end{subfigure}
    \begin{subfigure}[t]{\linewidth}
        \centering
        \begin{subfigure}[t]{.49\linewidth}
            \centering
            \includegraphics[width=\linewidth]{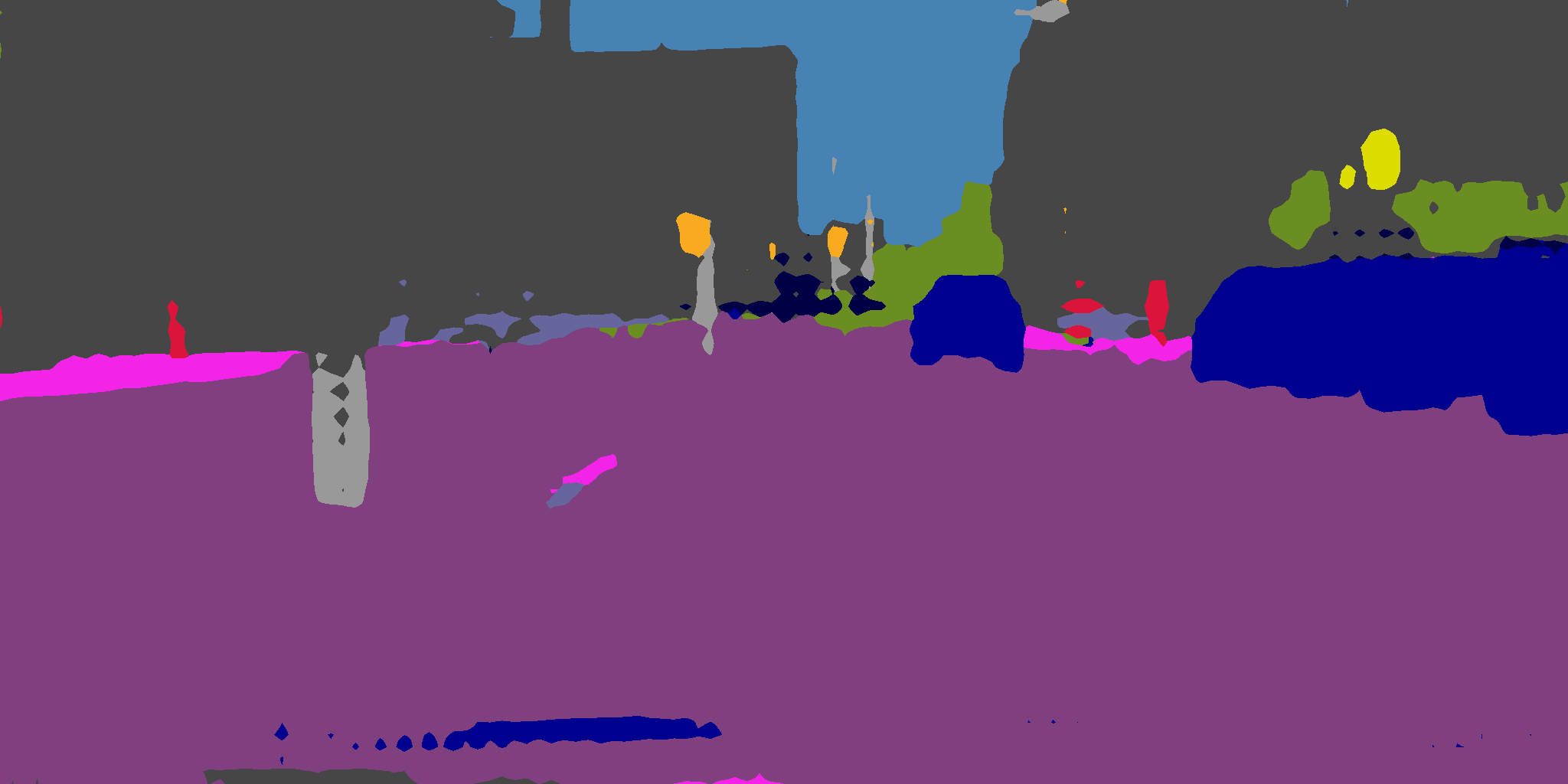}
        \end{subfigure}
        \begin{subfigure}[t]{.49\linewidth}
            \centering
            \includegraphics[width=\linewidth]{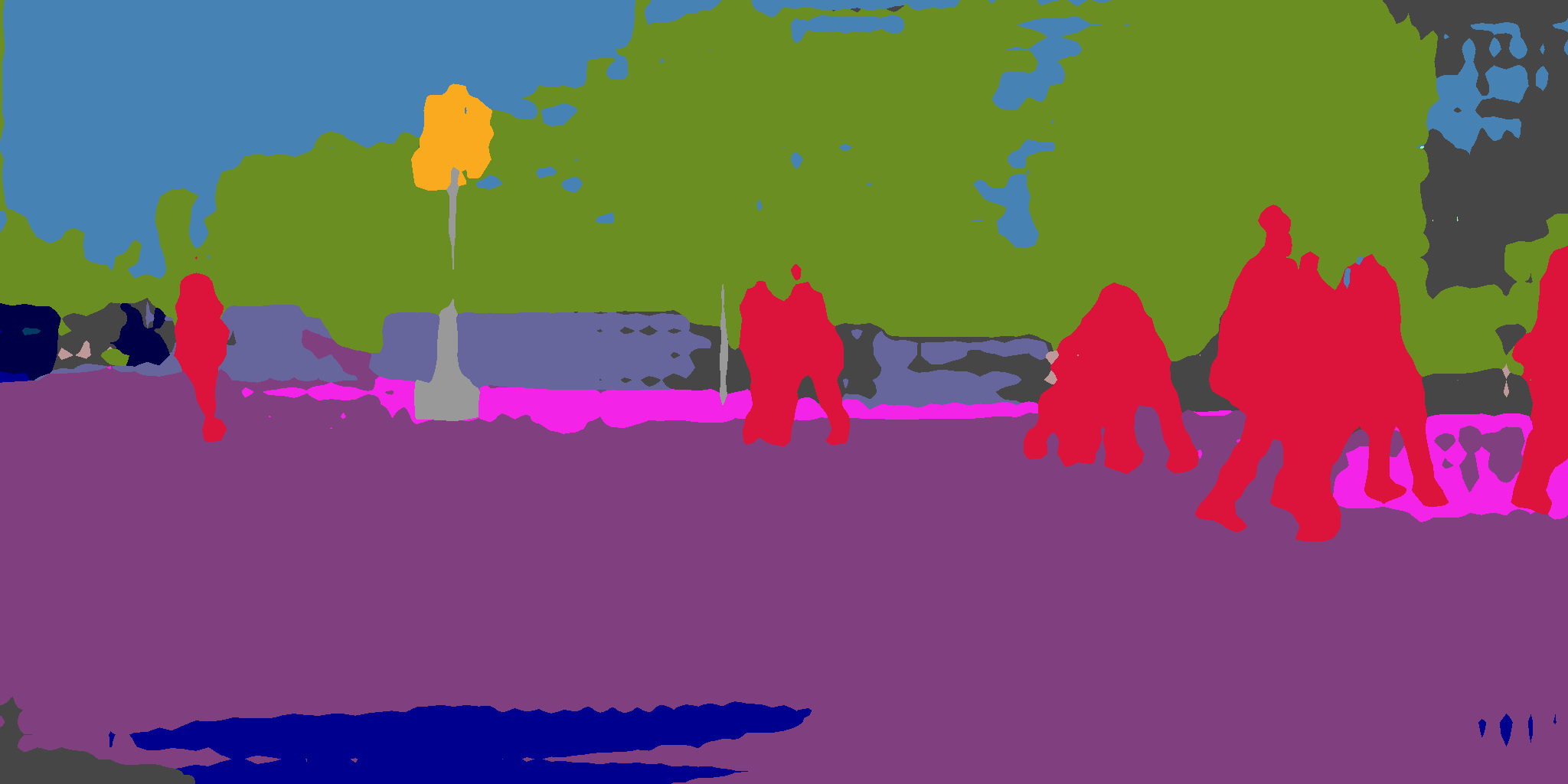}
        \end{subfigure}
        \caption{AdaptSegNet \cite{tsai_learning_2018}.}
        \vspace{5pt}
    \end{subfigure}
    \begin{subfigure}[t]{\linewidth}
        \centering
        \begin{subfigure}[t]{.49\linewidth}
            \centering
            \includegraphics[width=\linewidth]{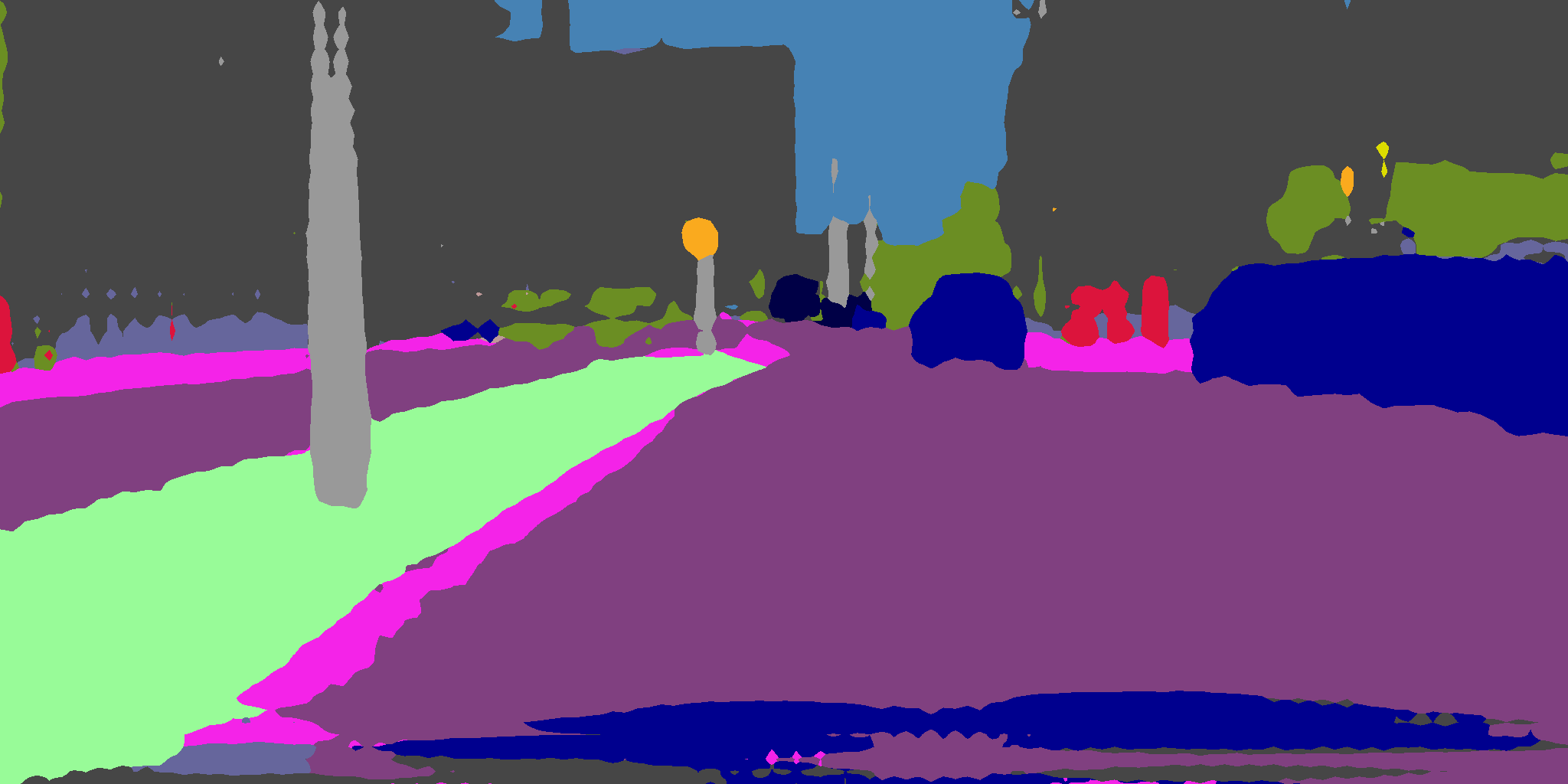}
        \end{subfigure}
        \begin{subfigure}[t]{.49\linewidth}
            \centering
            \includegraphics[width=\linewidth]{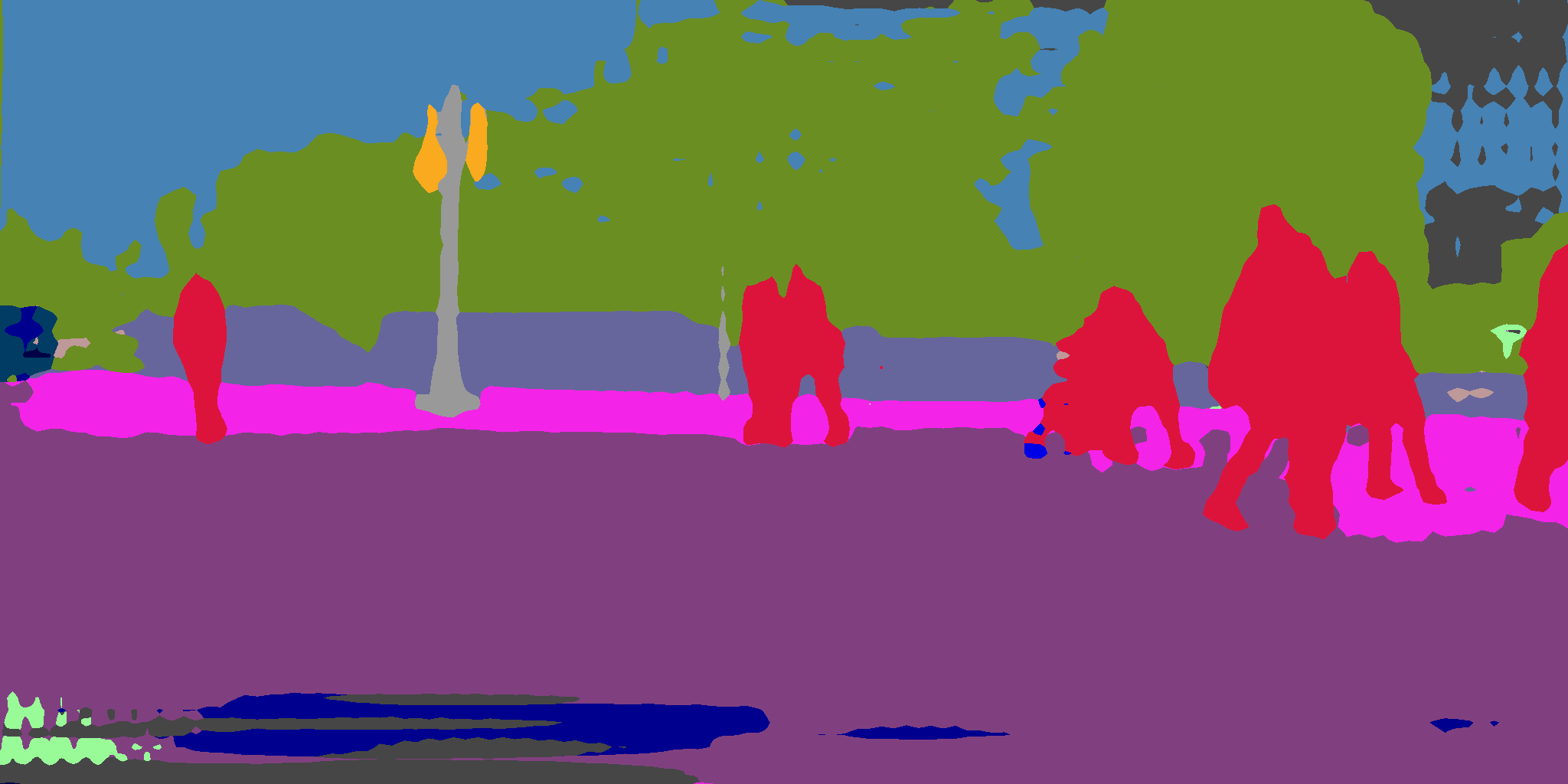}
        \end{subfigure}
        \caption{Our proposed unsupervised method based on AdaptSegNet \cite{tsai_learning_2018}.}
        \vspace{5pt}
    \end{subfigure}
        \begin{subfigure}[t]{\linewidth}
        \centering
        \begin{subfigure}[t]{.49\linewidth}
            \centering
            \includegraphics[width=\linewidth]{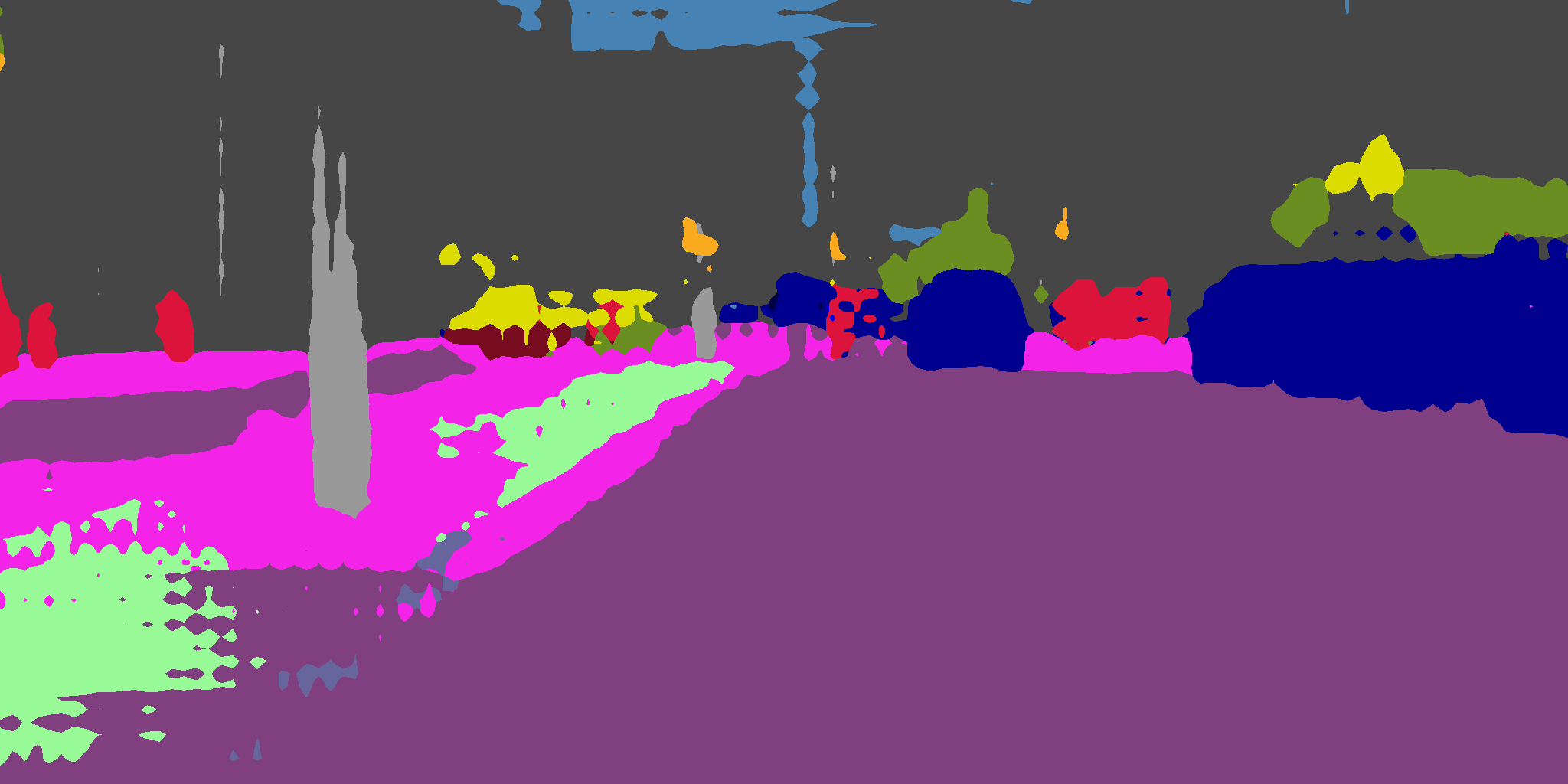}
        \end{subfigure}
        \begin{subfigure}[t]{.49\linewidth}
            \centering
            \includegraphics[width=\linewidth]{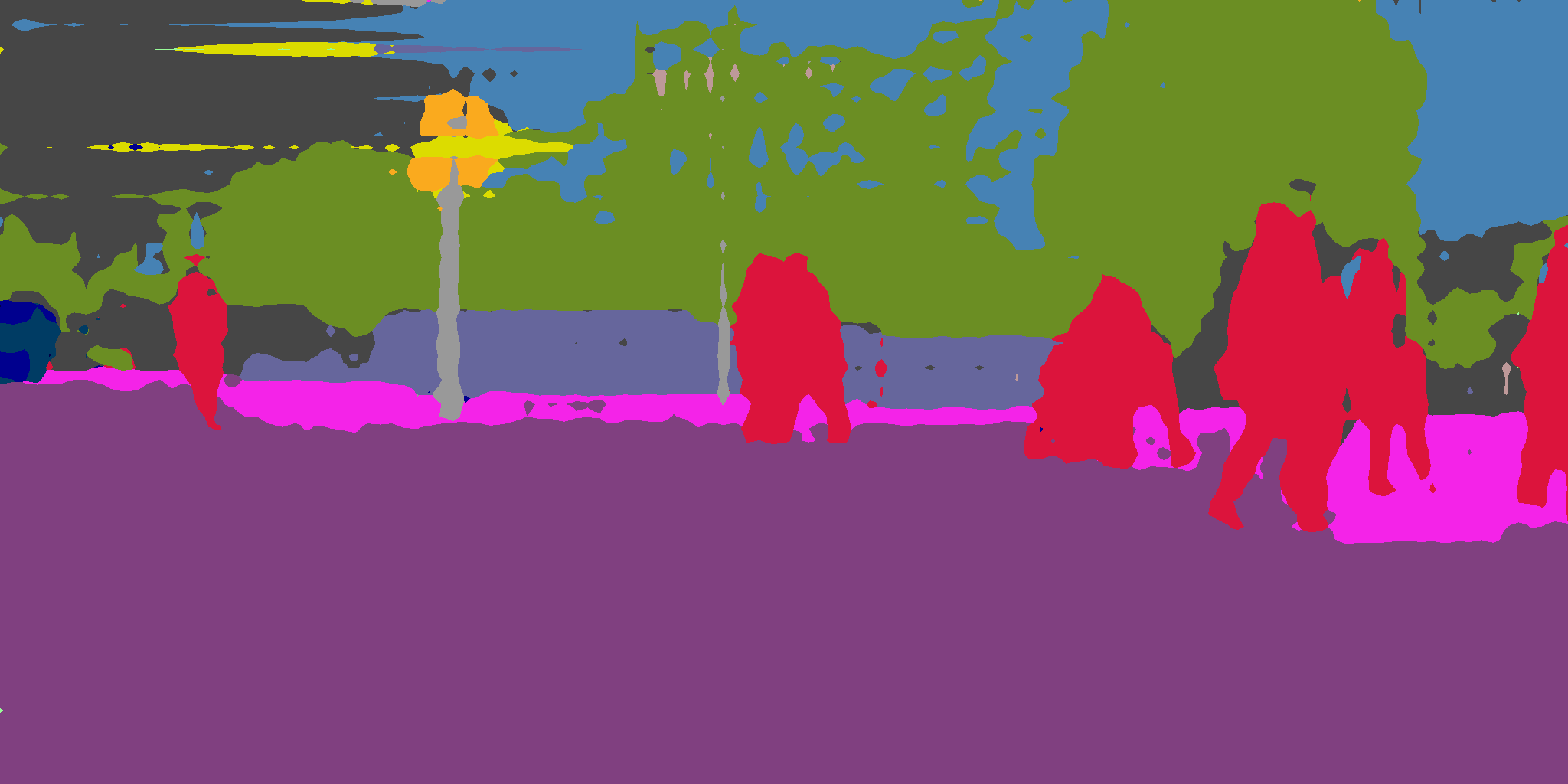}
        \end{subfigure}
        \caption{Conventional supervised domain adaptation.}
        \vspace{5pt}
    \end{subfigure}
        \begin{subfigure}[t]{\linewidth}
        \centering
        \begin{subfigure}[t]{.49\linewidth}
            \centering
            \includegraphics[width=\linewidth]{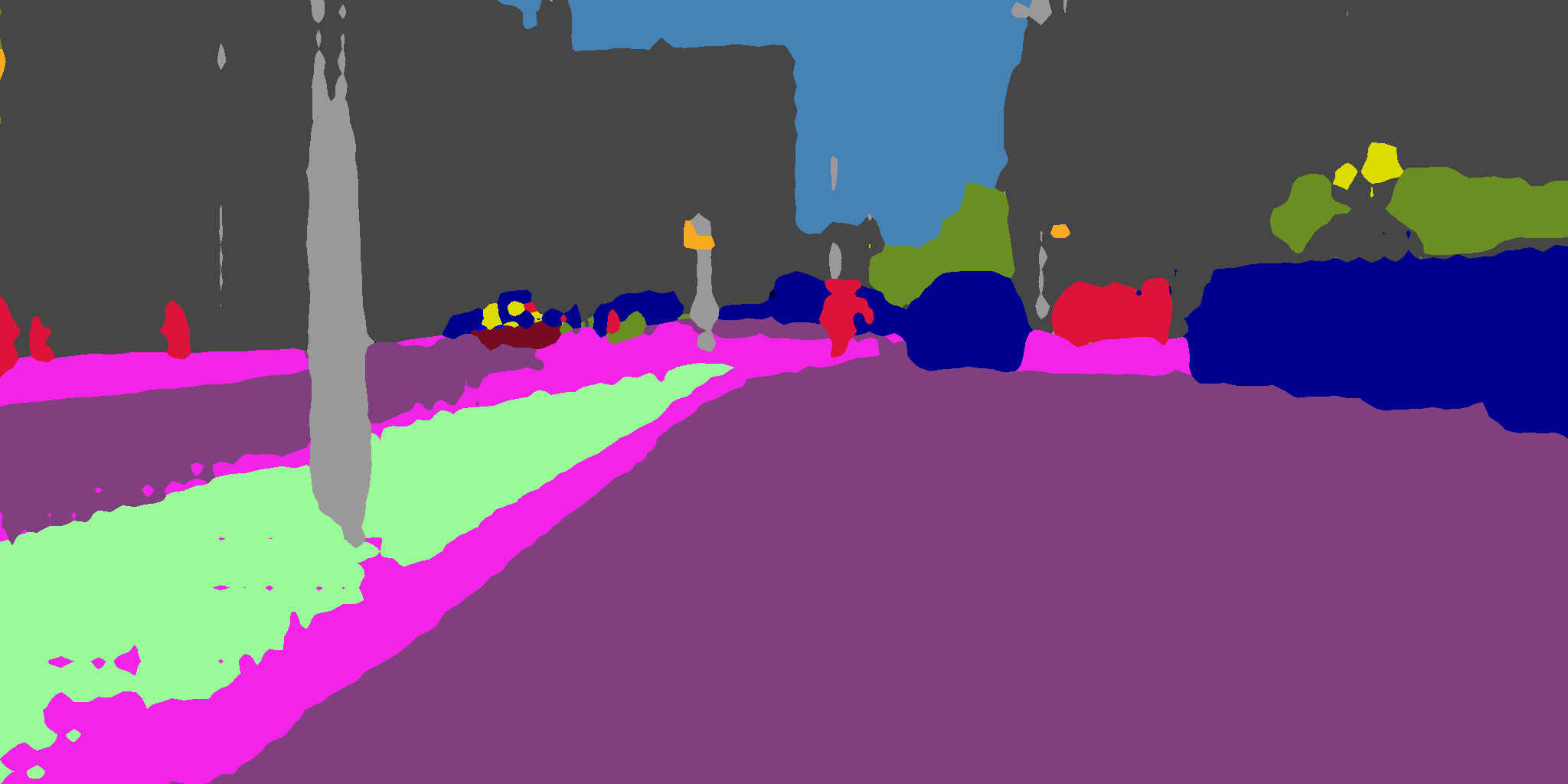}
        \end{subfigure}
        \begin{subfigure}[t]{.49\linewidth}
            \centering
            \includegraphics[width=\linewidth]{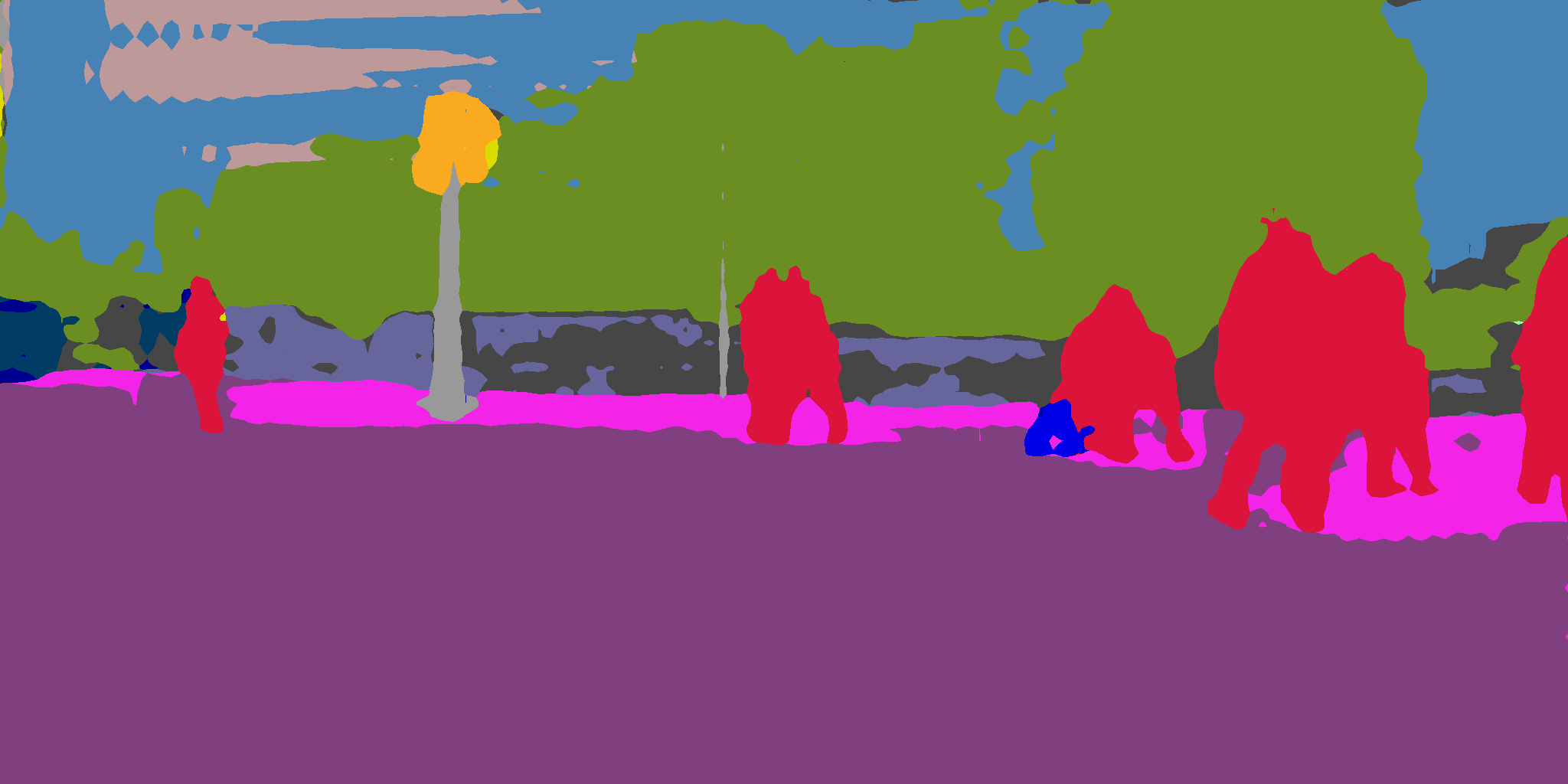}
        \end{subfigure}
        \caption{Our proposed supervised method.}
        \vspace{5pt}
    \end{subfigure}
    \begin{subfigure}[t]{\linewidth}
        \centering
        \begin{subfigure}[t]{.49\linewidth}
            \centering
            \includegraphics[width=\linewidth]{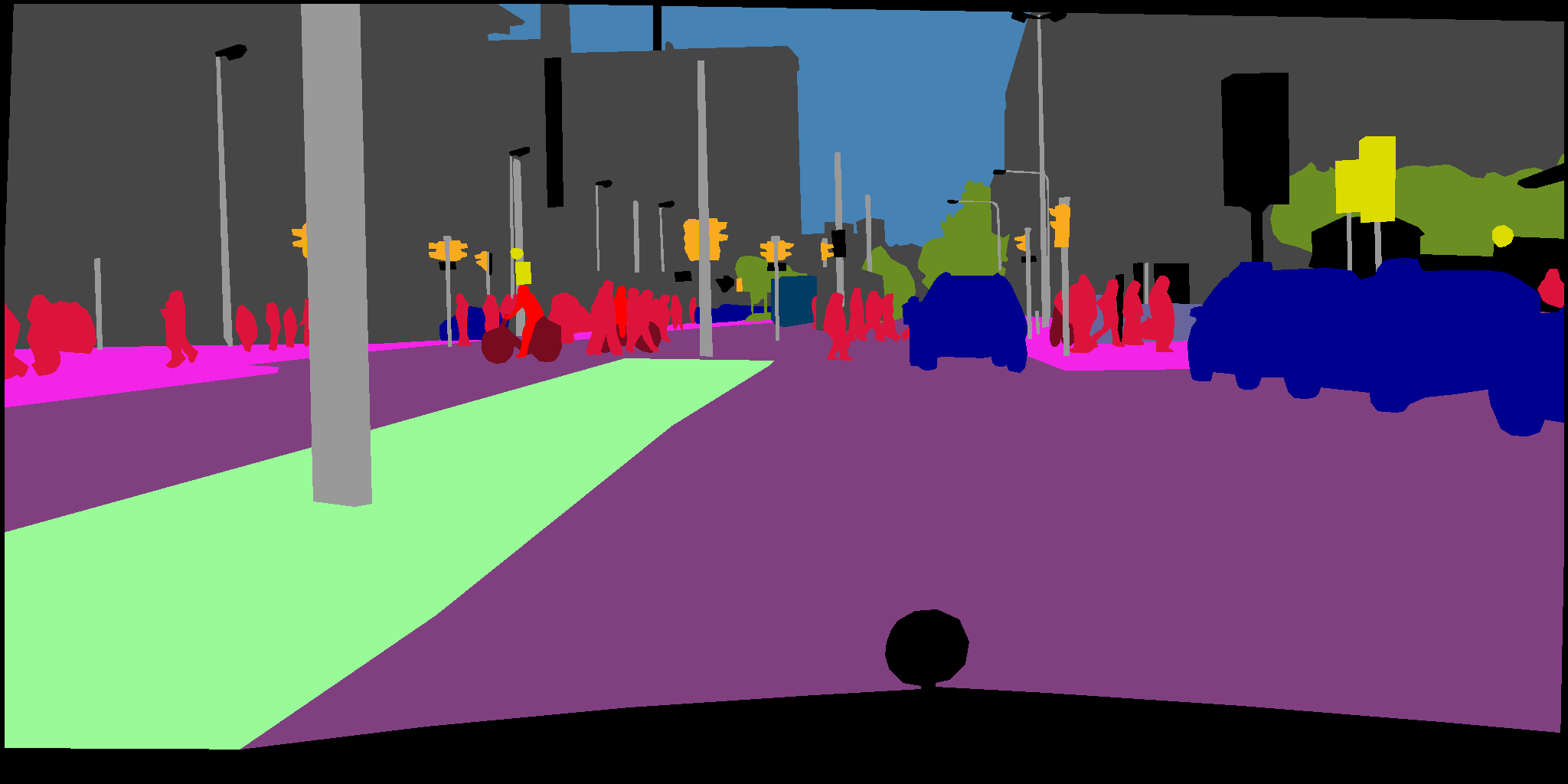}
        \end{subfigure}
        \begin{subfigure}[t]{.49\linewidth}
            \centering
            \includegraphics[width=\linewidth]{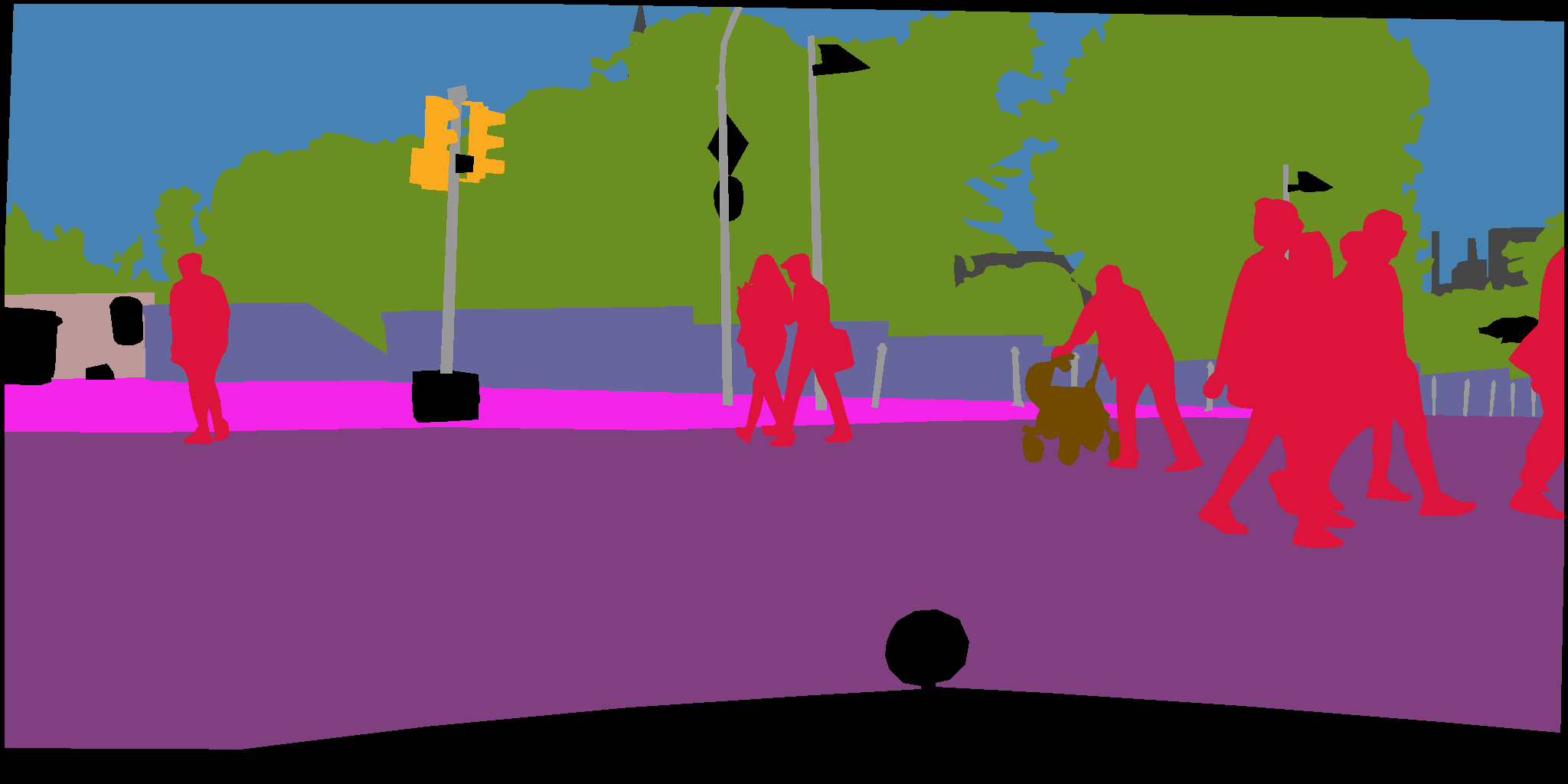}
        \end{subfigure}
        \caption{Groundtruth segmentation mask.}
    \end{subfigure}
    \caption{Comparison of qualitative results. All frameworks employ DeepLab-v2 with the ResNet-50 backbone.}
    \label{fig:example_images} \vspace{-2mm}
\end{figure} 
\subsection{Unsupervised Domain Adaptation}
For evaluation on the task of unsupervised domain adaptation, we compare the proposed domain adaptation method to the state of the art method by Tsai et al. \cite{tsai_learning_2018} on domain transfer from Playing for Data \cite{leibe_playing_2016} to Cityscapes \cite{cordts_cityscapes_2016}. As domain adaptation performance highly fluctuates between runs, which is due to the approaches being based on unstable adversarial networks, we perform ten runs with our method as well as the baseline network.

The generator network is trained with an initial learning rate of 0.00025 and polynomial decay. Training is stopped early after 160,000 iterations for the conventional training. Pre-training on the combined dataset is terminated after 85,000 iterations and continues with fine-tuning on the conventional dataset for additional 75.000 iterations.

Comparison of the average performance of both approaches, measured as the mean intersection over union (mIoU) on the validation set, is shown in Figure \ref{fig:unsupervised performance}. The blue plot corresponds to the conventional training procedure and reaches its maximum performance after 115,000 iterations with a mIoU of $33.18\,\%$, which is in line with numbers reported in literature for the ResNet-50 backbone. Results generated with the network pre-trained on the combined datasets are shown in red. After the switch to fine-tuning at iteration 85,000, the networks outperform the conventional approach at most steps. Optimal results are generated at iteration 150,000 with a mIoU of $34.20\,\%$. Furthermore, pre-training with the combined dataset outperforms the maximum of the conventional training, indicated by the horizontal dashed line, at 11 out of 15 evaluated steps.

While most previous work in the community reports the single best run, we believe that this does not well incorporates the use-case for unsupervised domain adaptation. As a great performance gain can be achieved even from very few annotated images in the target domain, fully unsupervised domain adaptation cannot rely on a validation dataset to select the best performing checkpoint. Averaging over multiple training runs reduces the impact of fluctuations, nevertheless, by evaluating not at a single iteration, but stating the average performance over a window further approximates the practical setting, where the optimal termination point cannot be determined exactly. We set the window length to seven time-steps and optimize its location for each method individually. For the conventional implementation, the optimal window location incorporates iterations 110,000 to 135,000, for the network pre-trained on the combined dataset, the optimal window starts at iteration 125,000 and ends with iteration 150,000. Using this approach, the overall and per class IoU for both methods is reported in Table~\ref{table:unsupervised results}.

\subsection{Supervised Domain Adaptation with Limited Data}
Fine-tuning on a small set of labeled images from the target domain is evaluated with subsets of 5, 10 and 20 images from Cityscapes, where the sets are defined such that they do not contain multiple images from a single city. Validation is performed on a disjoint set of 100 images, randomly selected from the original training set and the 500 Cityscapes validation images are employed as testset. The initial learning rate for fine-tuning is set to 0.00001 with polynomial learning rate decay and images are drawn at random from the small dataset.

We compare the conventional baseline, pre-trained on Playing for Data, with one network pre-trained on a stylized version and one network trained on the combination of both. Results on the test set, generated from the best performing checkpoints on the validation set, are reported in Table~\ref{table:supervised results}. Out of all training approaches, pre-training on combined Playing for Data performs best in all experiments.

\section{DISCUSSION AND CONCLUSION}
Our experiments confirm that texture underfitting can improve the performance of domain adaptation. In the unsupervised setting, we were able to show that joint training with stylized as well as conventional data can effectively enhance existing domain adaptation techniques. The results of our experiments on fine-tuning with limited data support the intuition that pre-training is especially relevant if only very few data points are available for training. The resulting mIoU from experiments with five images span a range of $6.81\,\%$ (from $32.84\,\%$ to $39.65\,\%$) depending on the pre-training, which shrinks to $3.32\,\%$ ($38.60\,\%$ to $41.92\,\%$) for 20 images. In the first case, while extensive training would lead to overfitting, short training still places a high weight onto the pre-trained network. In contrast to this, if more data is available for fine-tuning, longer training is possible, which results in less influence of the pre-training and consequently reduces the differences. In the presented use-case of supervised domain adaptation from synthetic to real (Playing for Data to Cityscapes), we identify the threshold at 20 images, where the performance of the conventional and our top performing setup approach each other.

While texture underfitting shows promising results, it needs to be employed thoughtfully. Even though envisioned differently, image stylization inevitably perturbs local structure. While this is not a problem for image classification and thus did not impact the results from Geirhos et al. \cite{geirhos_imagenet-trained_2018}, it impedes the image segmentation process as these structures often define object boundaries. The effect becomes especially apparent when the network is pre-trained on a stylized dataset only, which results in the degradation of its performance. Joint training on the stylized and the conventional dataset allows to circumvent the impact of this property.

Qualitative evaluation of the segmentation results as shown in Figure \ref{fig:example_images} reveals that training with the combined dataset mostly improves performance on well defined shapes like sidewalks or buildings. Furthermore, for both adaptation settings, our approach reduces artifacts where only a few pixels are segmented as incorrect classes, reducing the amount of well visible holes in the masks.

In order to further formalize the concept of texture underfitting for domain adaptation, it is interesting to have a solution by designing a new network architecture. This is left as future work.

\addtolength{\textheight}{-0cm}   % This command serves to balance the column lengths

%%%%%%%%%%%%%%%%%%%%%%%%%%%%%%%%%%%%%%%%%%%%%%%%%%%%%%%%%%%%%%%%%%%%%%%%%%%%%%%%

%%%%%%%%%%%%%%%%%%%%%%%%%%%%%%%%%%%%%%%%%%%%%%%%%%%%%%%%%%%%%%%%%%%%%%%%%%%%%%%%

%%%%%%%%%%%%%%%%%%%%%%%%%%%%%%%%%%%%%%%%%%%%%%%%%%%%%%%%%%%%%%%%%%%%%%%%%%%%%%%%
%\section*{APPENDIX}

%\section*{ACKNOWLEDGMENT}

%The preferred spelling of the word ÒacknowledgmentÓ in America is without an ÒeÓ after the ÒgÓ. Avoid the stilted expression, ÒOne of us (R. B. G.) thanks . . .Ó  Instead, try ÒR. B. G. thanksÓ. Put sponsor acknowledgments in the unnumbered footnote on the first page.

%%%%%%%%%%%%%%%%%%%%%%%%%%%%%%%%%%%%%%%%%%%%%%%%%%%%%%%%%%%%%%%%%%%%%%%%%%%%%%%%
\vspace{1mm}
\noindent
\textbf{Acknowledgement}: The work is supported by Toyota Motor Europe via the research project TRACE-Z\"urich.

%\bibliography{references.bib}
\bibliographystyle{./IEEEtranS}
\bibliography{ms}

%\begin{thebibliography}{99}
%\end{thebibliography}

\end{document}